\documentclass[letterpaper, 10 pt, conference]{ieeeconf}  

\IEEEoverridecommandlockouts                              

\usepackage{graphics} 
\usepackage{subfigure}
\usepackage{amsfonts}
\usepackage{amsmath,bm}
\usepackage{graphicx}

\usepackage{amsthm}
\usepackage[usenames,dvipsnames]{color} 

\usepackage{cite}

\title{\LARGE \bf
Towards Exact Interaction Force Control for Underactuated Quadrupedal Systems with Orthogonal Projection and Quadratic Programming
}

\author{Shengzhi Wang$^{1}$, Xiangyu Chu$^{1*}$, and K. W. Samuel Au$^{1}$ 
\thanks{$^{1}$The authors are with the Department of Mechanical and Automation Engineering at The Chinese University of Hong Kong, and with the Multiscale Medical Robotics Center, Hong Kong, China. $*$: Corresponding author.
        \tt\small \{shengzhiwang, xiangyuchu, samuelau\}@mrc-cuhk.com}%
}

\newtheorem{remark}{Remark}
\newtheorem{assumption}{Assumption}

\begin{document}

\maketitle
\thispagestyle{empty}
\pagestyle{empty}

\begin{abstract}

        Projected Inverse Dynamics Control (PIDC) is commonly used in robots subject to contact, especially in quadrupedal systems. Many methods based on such dynamics have been developed for quadrupedal locomotion tasks, and only a few works studied simple interactions between the robot and environment, such as pressing an E-stop button. To facilitate the interaction requiring exact force control for safety, we propose a novel interaction force control scheme for underactuated quadrupedal systems relying on projection techniques and Quadratic Programming (QP). This algorithm allows the robot to apply a desired interaction force to the environment without using force sensors while satisfying physical constraints and inducing minimal base motion. Unlike previous projection-based methods, the QP design uses two selection matrices in its hierarchical structure, facilitating the decoupling between force and motion control. The proposed algorithm is verified with a quadrupedal robot in a high-fidelity simulator. Compared to the QP designs without the strategy of using two selection matrices and the PIDC method for contact force control, our method provided more accurate contact force tracking performance with minimal base movement, paving the way to approach the exact interaction force control for underactuated quadrupedal systems.

\end{abstract}

\section{Introduction}





Quadrupedal systems, whose six Degree of Freedom (DoF) floating base is considered to be passively connected to an inertial frame, are generally underactuated. To control such systems for locomotion and manipulation in real-world deployment, \textit{underactuation} must be addressed while designing a whole-body controller.
A paradigm based on orthogonal projection and Quadratic Programming (QP) has been exploited.
Within it, a projection matrix stems from the contact between a foot or an additional arm and the environment, helping eliminate contact forces and thus reducing variables in optimization. Besides, the projection matrix can create two spaces: motion space and constraint space, allowing more freedom for task-oriented applications. QP can minimize a quadratic cost that resolves underactuation and accommodate constraints such as unilateral contact and friction cones.  This paradigm normally works for locomotion tasks or simple interaction tasks like pressing an emergency button \cite{XinOneLeg2020}. However, it is still open to using such a paradigm to achieve interaction applications that need exact force control. For example, as shown in Fig. 1, a robot uses one leg to keep pressing a button for allowing other robots to move into an elevator, and the robot can keep standing and have minimal base movement at the same time. 
In this case, exact force control is preferred for avoiding the robot's falling since impedance control for inducing force may generate external force disturbance due to an unknown environment.
Motivated by this need, in this paper, we focus on how the underactuated quadrupedal system applies an exact force to the environment subject to physical constraints and minimal disturbance on the robot's base.

\begin{figure}[t!]
            \includegraphics[width=0.485\textwidth, scale=1]{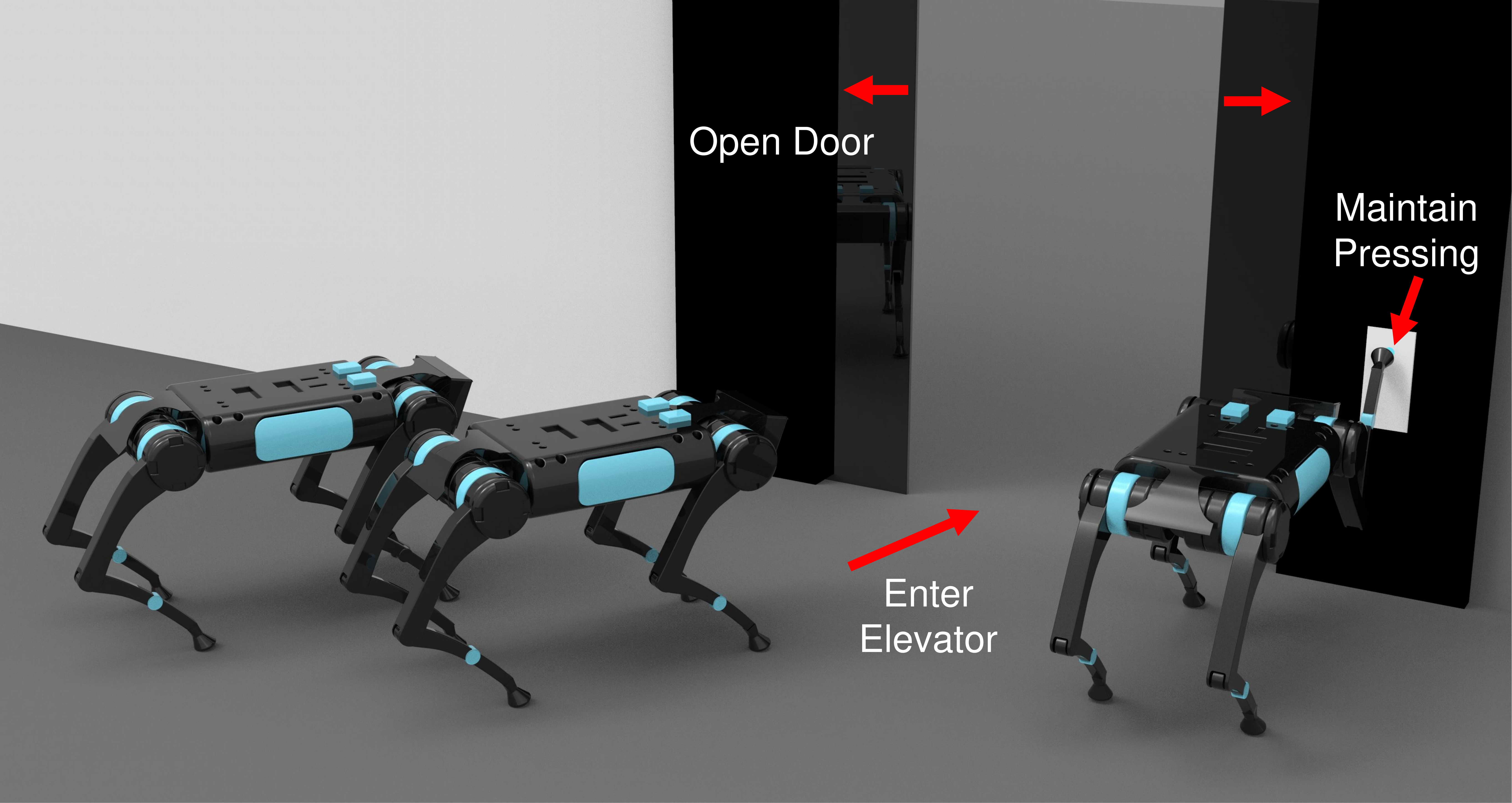}
            \caption{
            A scenario of a quadrupedal robot exerting desired force on the environment. Here, exact interaction force control of the raised leg is preferred, because, on the one hand, it can succeed to maintain pressing the button and keep the elevator door open, and on the other hand, the robot will keep standing without falling.
            }
            \label{fig:simulation}
\end{figure}


Projection-based methods have a long history. Projected Inverse Dynamics Control (PIDC) was first proposed for fully-actuated systems in \cite{Aghili2005}, paving the way to design controllers in motion and constraint space. Some researchers extended this idea to underactuated systems \cite{Mistry2010Floating, mistry2012operational,chu2021operational,xin2020optimization}. For example, the work \cite{mistry2012operational} used either null space motion or constraint forces to resolve underactuation without affecting task-space dynamics, while \cite{xin2020optimization} only made use of constraint forces. The aforementioned studies focused on motion control levels, and their constraint forces do not need to be specially considered in the controller. To impose more authority on the constraint force for underactuated applications, the constraint force is optimized within constraint space to maintain contact \cite{dehio2018modeling, dehio2022enabling}. For example, \cite{dehio2022enabling} designed an optimization problem to seek the optimal contact wrenches that minimize torques while satisfying physical constraints and compensating external forces. Although the constraint force/wrench was specially treated, those methods still cannot provide an exact contact force actively since the force/wrench was implicitly manipulated. Previously, explicitly tracking desired contact forces in quadrupedal systems was implemented in \cite{RighettiDistribution2013} but required planning on both desired position trajectories and desired contact forces at the same time if imposing higher priority on force control than motion control, because desired accelerations affected realizable contact forces. The tracking of \textit{planned} force profiles may not be suitable for tasks requiring fast reaction (e.g., tracking desired interaction force command from users); thus, a reactive control scheme is preferable.

To this end, to approach an exact force output, we propose a novel interaction force control scheme for underactuated quadrupedal systems in the sense of reaction control. It allows the system (e.g., using a foot) to apply a force as precisely as possible to the environment, without requiring force planning. Our scheme uses two QP designs that resolve the underactuation problem by splitting it into two orthogonal spaces: motion and constraint space, and then optimizing the cost in each space in a hierarchical order. Physical limitations (i.e., unilateral contact, torque limits, and friction cones) are also considered as inequality constraints in the design. To decouple force and motion control and accomplish the exact constraint force control as much as possible, we apply two selection matrices, in which one of them distributes the desired force control task to the designated joints, and another one selects the rest joints for the underlying motion task. For instance, as shown in Fig. \ref{fig:simulation}, a quadrupedal robot uses its front right leg for force control, while the other three legs support its base and conduct the motion task of the base. These two selection matrices select the front right leg and the other three legs for the force and motion control, respectively. In summary, our control scheme is a reactive control (e.g., can 
respond to user-defined force inputs fast) that does not rely on any motion planning techniques and force sensors (FS), and induces minimal base motion.

The contributions of this paper are:

\begin{itemize}
    \item[1)] Presenting a novel interaction force control scheme for underactuated quadrupedal systems that does not require motion planning and FS. 
    \item[2)] For resolving the underactuation problem, we propose a hierarchical QP structure that minimizes the cost function for the motion and force control in motion and constraint space, respectively. To reduce the coupling effect between motion and force control, two selection matrices are deployed, allowing us to decouple force and motion to the greatest extent and thus approach the exact force control. 
\end{itemize}







\section{Background}
\subsection{Projected Inverse Dynamics with External Disturbance}

\label{Projection Dynamics}
The dynamics equation of a quadruped robot can be expressed as:
\begin{equation}
        \label{dynEqa}
        \boldsymbol{M} \ddot{\boldsymbol{q}}+\boldsymbol{h} = \boldsymbol{B} \boldsymbol{\tau} + \boldsymbol{J}_{c}^{T} \boldsymbol{\lambda} + \boldsymbol{J}_{x}^{T} \boldsymbol{F}_x,
\end{equation}
where $\boldsymbol{q} = \left[\boldsymbol{q}_{b}^{T},\, \boldsymbol{q}_{j}^{T}\right]^{T}$ is the generalized coordinate vector including unactuated floating base coordinates $\boldsymbol{q}_{b} \in SE(3)$ and actuated joint configuration $\boldsymbol{q}_{j} \in \mathbb{R}^{n_j}$, $\boldsymbol{M} \in \mathbb{R}^{(6 + n_j) \times(6 + n_j)}$ denotes the inertia matrix, $\boldsymbol{h} \in \mathbb{R}^{(6 + n_j)}$ is the non-linear effect consisting of Coriolis, centrifugal and gravitational forces, $\boldsymbol{B} = \left[\boldsymbol{0}_{n_j \times 6},\, \boldsymbol{I}_{n_j \times n_j}\right]^{T}$ is the selection matrix\footnote{Unlike the selection matrix in previous papers, the selection matrix here is not  square.} of actuated joints, $\boldsymbol{\tau} \in \mathbb{R}^{n_j}$ is the actuated joint torques, $\boldsymbol{J}_{c} \in \mathbb{R}^{k \times(6 + n_j)}$ represents the \textit{constraint Jacobian} with $k = 3n_c$ ($n_c$ denotes the number of legs in contact with the environment and the contact is assumed as a point contact), $\boldsymbol{\lambda} \in \mathbb{R}^{k}$ denotes the generalized constraint force vector used to control unactuated $\boldsymbol{q}_{b}$, $\boldsymbol{J}_{x} \in \mathbb{R}^{n_e \times(6 + n_j)}$ is the Jacobian at $\boldsymbol{x}$, and $\boldsymbol{F}_x \in \mathbb{R}^{n_e}$ is the external disturbances due to the interaction from human or environments. 

During locomotion, the support feet should not slip, i.e., the constraint $\boldsymbol{J}_{c} \dot{\boldsymbol{q}} = \boldsymbol{0}$ must be satisfied. This constraint indicates that any admissible $\dot{\boldsymbol{q}}$ lies in the constraint null space $\mathcal{N}\left(\boldsymbol{J}_{c}\right)$. According to \cite{aghili2005unified}, the dynamics equation $\eqref{dynEqa}$ can be projected into two subspaces by using the orthogonal projection matrix $\boldsymbol{P}$ and $\boldsymbol{I} - \boldsymbol{P}$ respectively as:
\begin{equation}
        \label{dynInP}
        \boldsymbol{P M} \ddot{\boldsymbol{q}}+\boldsymbol{P h}=\underbrace{\boldsymbol{P B \tau}}_{\boldsymbol{\tau}_m} + \boldsymbol{P} \boldsymbol{J}_{x}^{T} \boldsymbol{F}_x,
\end{equation}
\begin{equation}
        \label{dynInIMinusP}
        (\boldsymbol{I}-\boldsymbol{P})(\boldsymbol{M} \ddot{\boldsymbol{q}}+\boldsymbol{h})=\underbrace{(\boldsymbol{I}-\boldsymbol{P}) \boldsymbol{B} \boldsymbol{\tau}}_{\boldsymbol{\tau}_c}+\boldsymbol{J}_{c}^{T} \boldsymbol{\lambda} + (\boldsymbol{I}-\boldsymbol{P}) \boldsymbol{J}_{x}^{T} \boldsymbol{F}_x,
\end{equation}
where $\boldsymbol{P} = \boldsymbol{I} - \boldsymbol{J}_{c}^+ \boldsymbol{J}_{c}$ implies that the orthogonal null space projection matrix is computed from the Moore-Penrose pseudoinverse of $\boldsymbol{J}_{c}$ and projects vectors into the null space of the constraint, such that $\boldsymbol{P} = \boldsymbol{P}^2 = \boldsymbol{P}^T$, $\boldsymbol{P} \boldsymbol{J}_c^T = \boldsymbol{0}$, and $\boldsymbol{P}\dot{\boldsymbol{q}} = \dot{\boldsymbol{q}}$ for all $\dot{\boldsymbol{q}} \in \mathcal{N}\left(\boldsymbol{J}_{c}\right)$. Then $\boldsymbol{I}-\boldsymbol{P}$ represents the complementary projection into $\mathcal{N}^{\perp}(\boldsymbol{J}_{c})$. 

To solve the equation of $\boldsymbol{\lambda}$, $\ddot{\boldsymbol{q}}$ must be computed first. However, $\ddot{\boldsymbol{q}}$ cannot be obtained directly through pre-multiplying the inverse of $\boldsymbol{PM}$ in $\eqref{dynInP}$, as $\boldsymbol{PM}$ cannot be inverted attributed to the rank deficiency of $\boldsymbol{P}$. With the help of additional equations $(\boldsymbol{I}-\boldsymbol{P})\dot{\boldsymbol{q}} = \boldsymbol{0}$ and its derivative $(\boldsymbol{I}-\boldsymbol{P})\ddot{\boldsymbol{q}} = \dot{\boldsymbol{P}} \dot{\boldsymbol{q}}$ that are derived from $\boldsymbol{P}\dot{\boldsymbol{q}} = \dot{\boldsymbol{q}}$, $\ddot{\boldsymbol{q}}$ can be solved by substituting the latter equation into $\eqref{dynInP}$ as:
\begin{equation}
\label{eqn::ddq}
 \ddot{\boldsymbol{q}}=\boldsymbol{M}_{c}^{-1} (\boldsymbol{\tau}_m - \boldsymbol{P h} +\dot{\boldsymbol{P}} \dot{\boldsymbol{q}} + \boldsymbol{P} \boldsymbol{J}_x^T \boldsymbol{F}_{x} ), 
\end{equation}
where $\boldsymbol{M}_{c} = \boldsymbol{P M}+\boldsymbol{I}-\boldsymbol{P}$. Equations \eqref{eqn::ddq} shows that only the motion torques $\boldsymbol{\tau}_m$ contributes to the motion of the system, therefore $\mathcal{N}\left(\boldsymbol{J}_{c}\right)$ is called the \textit{motion space} as described in \cite{xin2020optimization}. Similarly, because in $\eqref{dynInIMinusP}$ the $\ddot{\boldsymbol{q}}$ is fixed and constraint forces $\boldsymbol{\lambda}$ are free to choose depending on the constraint torques $\boldsymbol{\tau}_c$, $\mathcal{N}^{\perp}(\boldsymbol{J}_{c})$ is named the \textit{constraint space}. Eventually, the constraint forces are obtained by inserting $\ddot{\boldsymbol{q}}$ into $\eqref{dynInIMinusP}$ and yields:
\begin{equation}
\begin{aligned}
        \label{constraint_force}
         \boldsymbol{\lambda}=(\boldsymbol{J}_{c}^{T})^+\Big[&(\boldsymbol{I}-\boldsymbol{P})[\bar{\boldsymbol{M}}(\boldsymbol{\tau}_m - \boldsymbol{P h} + \dot{\boldsymbol{P}} \dot{\boldsymbol{q}}) + \boldsymbol{h}] - \boldsymbol{\tau}_c \\
         &+ (\boldsymbol{I}-\boldsymbol{P})(\bar{\boldsymbol{M}} \boldsymbol{P} - \boldsymbol{I})\boldsymbol{J}_{x}^{T} \boldsymbol{F}_x \Big],
\end{aligned}
\end{equation}
with $\bar{\boldsymbol{M}} = \boldsymbol{M} \boldsymbol{M}_{c}^{-1}$, and  is the \textit{constraint inertia matrix} and is always invertible \cite{aghili2005unified}. More details can be found in \cite{xin2020optimization}.

\subsection{Cartesian Impedance Control in Task-Space}
\label{Cartesian Impedance Control}
To achieve the desired locomotion behavior, a Cartesian impedance controller is applied to the robot. We start with a closed-loop system equation that reflects a mechanical impedance of the end-effector in multi-dimension to the external disturbance $\boldsymbol{F}_x$, as:
\begin{equation}
        \label{impedance_response}
        \boldsymbol{\Lambda}_{d} \ddot{\boldsymbol{e}}+\boldsymbol{K}_{d} \dot{\boldsymbol{e}}+\boldsymbol{K}_{p} \boldsymbol{e}=\boldsymbol{F}_{x},
\end{equation}
where $\boldsymbol{e} = \boldsymbol{x} - \boldsymbol{x}_d$ denotes the pose error of the end-effector between the current and desired one, $\boldsymbol{\Lambda}_{d}$, $\boldsymbol{K}_{d}$, and $\boldsymbol{K}_{p}$ are the desired task-space inertia, damping, and stiffness matrices. Note that the torso and swing feet can be treated as the end-effector during the controller design. Again, we apply the equation $(\boldsymbol{I}-\boldsymbol{P})\ddot{\boldsymbol{q}} = \dot{\boldsymbol{P}} \dot{\boldsymbol{q}}$, as we compute the expression of $\ddot{\boldsymbol{q}}$ in Section \ref{Projection Dynamics}, to equation \eqref{dynInP}, then pre-multiply the resultant equation by $\boldsymbol{J}_x \boldsymbol{M}_c^{-1}$, replace $\boldsymbol{J}_x \ddot{\boldsymbol{q}}$ with $\ddot{\boldsymbol{x}} - \dot{\boldsymbol{J}_x} \dot{\boldsymbol{q}}$, and pre-multiply it by the \textit{operational space inertia matrix} $\boldsymbol{\Lambda}_{c} = (\boldsymbol{J}_x \boldsymbol{M}_c^{-1} \boldsymbol{P} \boldsymbol{J}_x^T)^{-1}$, the equation becomes:
\begin{equation}
        \label{unsimplified modified projection dynamics}
        \boldsymbol{\Lambda}_{c} \ddot{\boldsymbol{x}}+\boldsymbol{h}_{c} = \boldsymbol{\Lambda}_{c} \boldsymbol{J}_{x} \boldsymbol{M}_{c}^{-1} \boldsymbol{P B} \boldsymbol{\tau} + \boldsymbol{F}_{x},
\end{equation}
where $\boldsymbol{J}_x$ is the \textit{task Jacobian} defined by $\dot{\boldsymbol{x}} = \boldsymbol{J}_x \dot{\boldsymbol{q}}$, and $\boldsymbol{h}_c = \boldsymbol{\Lambda}_{c} \boldsymbol{J}_{x} \boldsymbol{M}_{c}^{-1}(\boldsymbol{P h}-\dot{\boldsymbol{P}} \dot{\boldsymbol{q}})-\boldsymbol{\Lambda}_{c} \dot{\boldsymbol{J}}_{x} \dot{\boldsymbol{q}}$ is the \textit{operational space non-linear effect}. At this moment, we consider the robot is \textit{fully actuated}, such that:
\begin{equation}
    \boldsymbol{PB \tau} = \boldsymbol{P}\boldsymbol{J}_x ^T \boldsymbol{F}_{d, x} = \boldsymbol{\tau}_{m, d},
\end{equation}
with $\boldsymbol{F}_{d, x}$ is the designed control law, and $\boldsymbol{\tau}_{m, d}$ is the desired motion torques. Then, equation $\eqref{unsimplified modified projection dynamics}$ will be simplified as:
\begin{equation}
        \label{simplified modified projection dynamics}
        \boldsymbol{\Lambda}_{c} \ddot{\boldsymbol{x}}+\boldsymbol{h}_{c} = \boldsymbol{F}_{d, x} + \boldsymbol{F}_{x}.
\end{equation}
The control law $\boldsymbol{F}_{d, x}$ can be defined to achieve the impedance response of equation $\eqref{impedance_response}$ as:
\begin{equation}
        \label{control_law}
        \boldsymbol{F}_{d, x}=\boldsymbol{h}_{c}+\boldsymbol{\Lambda}_{c} \ddot{\boldsymbol{x}}_{d}-\boldsymbol{K}_{d} \dot{\boldsymbol{e}}-\boldsymbol{K}_{p} \boldsymbol{e}, 
\end{equation}
where $\boldsymbol{\Lambda}_{d} = \boldsymbol{\Lambda}_{c}$ is assigned \cite{ott2008cartesian}. 
\begin{remark}
Note that $\boldsymbol{\Lambda}_{c}$ is not always determinable, since $\boldsymbol{J}_x \boldsymbol{M}_c^{-1} \boldsymbol{P} \boldsymbol{J}_x^T$ is not invertible when $\boldsymbol{J}_x$ has rank deficiency. We use Singular Value Decomposition (SVD) to remove the eigenvalues equal or close to zero for the approximation of $\boldsymbol{\Lambda}_{c}$.  
\end{remark}

\begin{remark}
Consider the torso and swing feet as end-effectors, all $\boldsymbol{J}_x$ above can be replaced by the Jacobian matrix of the torso $\boldsymbol{J}_b \in \mathbb{R}^{6 \times(6 + n_j)}$ and of the $i$-th foot $\boldsymbol{J}_i \in \mathbb{R}^{3 \times(6 + n_j)}$, and the control law $\boldsymbol{F}_{d, x}$ can be represented by $\boldsymbol{F}_{d, b} \in \mathbb{R}^{6}$ for the torso and $\boldsymbol{F}_{d, i} \in \mathbb{R}^{3}$ for the foot. 
\end{remark}

\subsection{Contact Force Control in Constraint Space}
\label{Sec::Contact Force Control in Constraint Space}

\begin{assumption}
The desired force $\boldsymbol{\lambda}_d$ is selected within a feasible force region that will not cause the robot to fall. Determining the feasible force region is out of the scope of this paper.  
\end{assumption}

As described in Section \ref{Projection Dynamics}, once $\ddot{\boldsymbol{q}}$ is determined, the contact forces can be controlled by applying corresponding constraint torques. This has been shown in \cite{ortenzi2014experimental, ortenzi2015projected}, where the robot executed a desired motion and applied a desired force at the contact point simultaneously. Here, we extend the constraint torques control law $\boldsymbol{\tau}_{c, d}$ with the external disturbance by:
\begin{equation}
\begin{aligned}
\label{eqn::tau_c_d_law}
\boldsymbol{\tau}_{c, d} = (\boldsymbol{I}-\boldsymbol{P})[&\bar{\boldsymbol{M}}(\boldsymbol{\tau}_{m, d} - \boldsymbol{P h} + \dot{\boldsymbol{P}} \dot{\boldsymbol{q}} + \boldsymbol{P} \boldsymbol{J}_x^T \boldsymbol{F}_{x})+ \boldsymbol{h}]\\
&- (\boldsymbol{I}-\boldsymbol{P}) \boldsymbol{J}_{x}^{T} \boldsymbol{F}_x - \boldsymbol{J}_c^T \boldsymbol{\lambda}_d.
\end{aligned}
\end{equation}

\begin{remark}
The external force $\boldsymbol{F}_x$ can be estimated by equation \eqref{impedance_response}. In this case, the usage of the FS can be avoided. 
\end{remark}
The complete torque commands that can accomplish the desired motion and contact force are the sum of $\boldsymbol{\tau}_{m, d}$ and $\boldsymbol{\tau}_{c, d}$, i.e.:
\begin{equation}
    \label{eqn::total_torque_command_background}
    \boldsymbol{B} \boldsymbol{\tau} = \boldsymbol{\tau}_{d} = \boldsymbol{\tau}_{m, d} + \boldsymbol{\tau}_{c, d}
\end{equation}
\begin{remark}
If the contact force control is not concerned, $\boldsymbol{\tau}_{c, d}$ can be set to zero vector. In this case, the contact forces are controlled implicitly\footnote{This differs from the classical implicit force control.}, since from \eqref{eqn::tau_c_d_law} $\boldsymbol{\tau}_{c, d} = \boldsymbol{0}$ implies:
\begin{equation}
\begin{aligned}
    \boldsymbol{\lambda}_d = (\boldsymbol{J}_c^T)^+ (\boldsymbol{I}-\boldsymbol{P})\Big[&[\bar{\boldsymbol{M}}(\boldsymbol{\tau}_{m, d} - \boldsymbol{P h} + \dot{\boldsymbol{P}} \dot{\boldsymbol{q}} + \boldsymbol{P} \boldsymbol{J}_x^T \boldsymbol{F}_{x}) \\
    &+ \boldsymbol{h}] - \boldsymbol{J}_{x}^{T} \boldsymbol{F}_x\Big],
\end{aligned}
\end{equation}
which are the contact forces (or in general, the constraint forces) needed to remain the constraint $\boldsymbol{J}_{c} \dot{\boldsymbol{q}} = \boldsymbol{0}$. It reveals that when the PIDC method is considered for pure motion control, the contact forces required to satisfy the constraint are inherently fulfilled. This virtue has rarely been studied and discussed in the prior literature. 
\end{remark}

\section{
Proposed Control Scheme
}
\begin{figure}[t!]
            \includegraphics[width=0.485\textwidth]{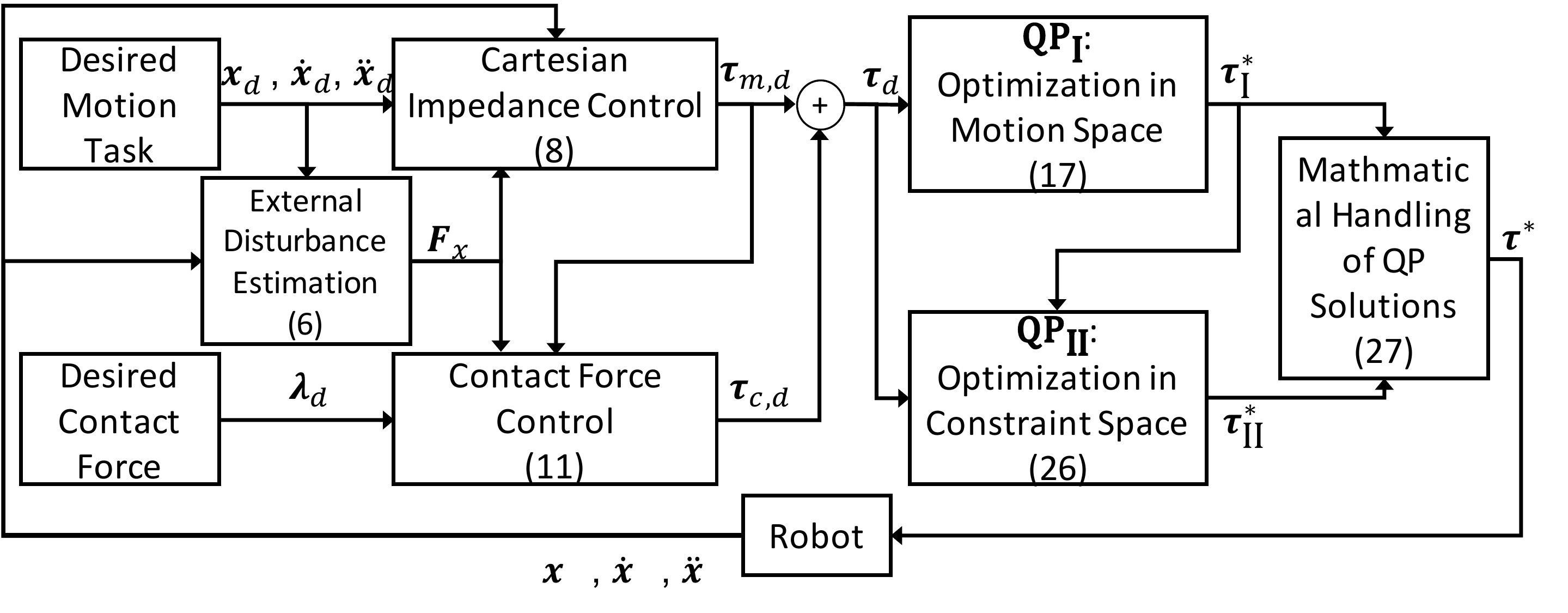}
            \caption{Our proposed controller schematic.}
            \label{fig:controller_schematic}
\end{figure}

The schematic of our proposed controller can be seen in Fig. \ref{fig:controller_schematic}. 
\subsection{Motivation}
\label{sec::motivation_in_Propose Contact Force Control with Underactuation}
In Section \ref{Cartesian Impedance Control} and \ref{Sec::Contact Force Control in Constraint Space}, the systems are assumed to be \textit{fully actuated} when deriving the desired motion and constraint torque commands. However, $\boldsymbol{B \tau} \neq \boldsymbol{\tau}_{d}$ in general for underactuated robots. To satisfy the underactuation constraint $\boldsymbol{B \tau} = \boldsymbol{\tau}_{d}$, \cite{dehio2018modeling} extended the PIDC formulation for the underactuated systems to realize the contact-consistent motion and contact force control by adding an additional constraint torques $(\boldsymbol{I - P})\boldsymbol{\tau}_u$ to equation \eqref{eqn::total_torque_command_background} as:
\begin{equation}
\begin{aligned}
\label{eqn::Contact-consistent motion generation and contact wrench control for the underactuation}
   \boldsymbol{\tau}_d &= \boldsymbol{\tau}_{m, d} + \boldsymbol{\tau}_{c, d} + (\boldsymbol{I - P})\boldsymbol{\tau}_u\\
   &= [\boldsymbol{P B}]^{+} \boldsymbol{\tau}_{m, d} + \left[\boldsymbol{I}-[(\boldsymbol{I}-\boldsymbol{B})(\boldsymbol{I}-\boldsymbol{P})]^{+}\right](\boldsymbol{I}-\boldsymbol{P}) \boldsymbol{\tau}_{c, d}.
\end{aligned}
\end{equation}
Indeed, the additional constraint torques $(\boldsymbol{I - P})\boldsymbol{\tau}_u$ does not generate any motion to violate the underlying task, but it derails the force control from its desired value, as $\boldsymbol{\tau}_{c, d}$ and $(\boldsymbol{I - P})\boldsymbol{\tau}_u$ are both in the constraint space. In other words, this approach fails to achieve exact force control even at any of all contact points. 

Exact interaction force control is an essential technique to accomplish many real tasks, e.g., grinding, polishing, and screwing. This control problem has been widely studied for fully actuated robots. Nevertheless, our work focuses on extending the PIDC framework for the underactuated systems, specifically, the floating base quadrupedal robots, towards exerting an accurate desired active force at the contact point.

Compared to the methods used in \cite{xin2020optimization, mistry2012operational, dehio2018modeling}, we aim to directly find an optimal solution that solves the optimization problem  $\min \left\|\boldsymbol{B \tau} - \boldsymbol{\tau}_{d}\right\|^2$ without inducing any additional null space components or constraint forces. However, because of the underactuation as mentioned in \cite{chu2021operational, braun2019operational}, the desired motion and contact forces cannot be implemented by directly solving this optimization problem. Fortunately, \cite{braun2019operational} offers an idea of \textit{constrained hierarchical optimization} that orthogonally decomposes the error norm into two subspaces at first, 
\begin{equation}
\label{eqn::error_norm_orthogonally_decomposed}
    \left\|\boldsymbol{B \tau} - \boldsymbol{\tau}_{d}\right\|^2 = \left\|\boldsymbol{B \tau} - \boldsymbol{\tau}_{d}\right\|^2_{\boldsymbol{P}} + \left\|\boldsymbol{B \tau} - \boldsymbol{\tau}_{d}\right\|^2_{\boldsymbol{I} - \boldsymbol{P}},
\end{equation}
where $\left\|\boldsymbol{B \tau} - \boldsymbol{\tau}_{d}\right\|^2_{\boldsymbol{P}}$ and $\left\|\boldsymbol{B \tau} - \boldsymbol{\tau}_{d}\right\|^2_{\boldsymbol{I} - \boldsymbol{P}}$ are the motion space and constraint space error norm, respectively. Then, the optimization problem of \eqref{eqn::error_norm_orthogonally_decomposed} will be solved hierarchically in the motion and constraint space. Following the constrained hierarchical optimization formulation, the rest of this section outlines our PIDC-based method for the contact force control of the underactuated robots without using FS at the contact point. 

\subsection{Motion Control for Underlying Task}
\label{Sec::Motion Control for Underlying Task}
As stated in \cite{aghili2005unified}, the constraint forces $\boldsymbol{\lambda}$ are affected by the constraint torques $\boldsymbol{\tau}_c$ directly and motion torques $\boldsymbol{\tau}_m$ indirectly via a cross-coupling factor $\mu = (\boldsymbol{I} - \boldsymbol{P})\bar{\boldsymbol{M}}$. It is apparent that $\boldsymbol{\tau}_m$ must be determined first, otherwise determining $\boldsymbol{\tau}_m$ afterward would affect the contact force control. Similar ideas have also been seen in \cite{ortenzi2014experimental, ortenzi2015projected, dehio2018modeling, lin2018projected, dehio2022enabling}. Therefore, the motion space error norm is minimized at first. 

The selection matrix $\boldsymbol{B}$ can be decomposed into two selection matrices $\boldsymbol{B} = \boldsymbol{B}_m + \boldsymbol{B}_f$, where $\boldsymbol{B}_m$ selects the actuators to perform the desired underlying motion, and $\boldsymbol{B}_f$ selects the actuators for the desired contact force control. An intuitive instance would be a quadruped robot using its front right leg to impose a desired contact force on the button while using the other legs to hold the desired torso pose, as shown in Fig. \ref{fig:simulation}. In this case, $\boldsymbol{B}_f$ activates the front right leg to exert the desired interaction force, whereas $\boldsymbol{B}_m$ activates the remaining legs for the motion task. This decomposition of the selection matrix can also be generalized to other underactuation systems, e.g., collaborative manipulation of the multi-robot team on a free-floating object \cite{dehio2018modeling, lin2018projected, dehio2022enabling}. Based on the selection matrices design, \eqref{eqn::error_norm_orthogonally_decomposed} can be further decomposed into:
\begin{equation}
\label{eqn::error_norm_orthogonally_decomposed_with_selection_matrix}
\begin{aligned}
    &\left\|\boldsymbol{B \tau} - \boldsymbol{\tau}_{d}\right\|^2_{\boldsymbol{P}} + \left\|\boldsymbol{B \tau} - \boldsymbol{\tau}_{d}\right\|^2_{\boldsymbol{I} - \boldsymbol{P}} \\
    = &\|\boldsymbol{B}_m \boldsymbol{\tau} + \boldsymbol{B}_f \boldsymbol{\tau} - \boldsymbol{\tau}_{d}\|^2_{\boldsymbol{P}} + \|\boldsymbol{B}_m \boldsymbol{\tau} + \boldsymbol{B}_f \boldsymbol{\tau} - \boldsymbol{\tau}_{d}\|^2_{\boldsymbol{I} - \boldsymbol{P}}
\end{aligned}
\end{equation}
where $\boldsymbol{B}_f \boldsymbol{\tau}$ in motion space and $\boldsymbol{B}_m \boldsymbol{\tau}$ in constraint space are set to zero vector, respectively, because we do not expect that the joints used for the force control contribute to the motion task, and vice versa. Thus, the optimization problem in motion space becomes:
\begin{subequations}
\begin{equation}
\label{eqn::QP_1}
    \mathrm{\textbf{QP}}_{\mathrm{\textbf{I}}}: \boldsymbol{\tau}_{\mathrm{I}}^{*}=\underset{\boldsymbol{\tau}}{\arg \min }\left\|\boldsymbol{B}_m \boldsymbol{\tau}-\boldsymbol{\tau}_{d}\right\|_{\boldsymbol{P}}^{2}, 
\end{equation}
\begin{equation}
\label{eqn::torque_limit_in_motion_space}
\text{s. t. \,\,\,} \boldsymbol{\tau}_{\text{min}} \leq \boldsymbol{B}^T\boldsymbol{P}\boldsymbol{B}_m \boldsymbol{\tau} \leq \boldsymbol{\tau}_{\text{max}}, 
\end{equation}
\end{subequations}
where \eqref{eqn::torque_limit_in_motion_space} is the actuation torque limits of the robot. Here, the pre-multiplication of $\boldsymbol{B}^T$ is to extract the actuated joints from the vertical concatenation vector of all the underactuated and actuated joints, because it is not a square matrix. Note the other physical constraints including unilateral contact and friction cones are considered within the optimization in the constraint space.

\subsection{Contact Force Control}
Since the motion space error norm has been minimized, we need to optimize the constraint space error norm subject to the physical constraints. 
\subsubsection{Modeling of Unilateral and Friction Cone Constraint}
Assume $\boldsymbol{\lambda}_i \in \mathbb{R}^{3}$ is a contact force at the $i$-th contact point, $\boldsymbol{n}_{x, i}$, $\boldsymbol{n}_{y, i}$, $\boldsymbol{n}_{z, i} \in \mathbb{R}^{3}$ are the heading, lateral, and normal vector of the contact surface, and $\mu_i$ is the friction coefficient. The unilateral and friction cone constraint at this point is:
\begin{equation}
\label{eqn::unilateral_and_friction_cone_constraint}
   \underbrace{\left[\begin{array}{c}-\infty \\ -\infty \\ 0 \\ 0 \\ 0 \end{array}\right]}_{\underline{\boldsymbol{d}}} \leq \underbrace{\left[\begin{array}{c}(\boldsymbol{n}_{x, i} - \mu_i \boldsymbol{n}_{z, i})^T \\ (\boldsymbol{n}_{y, i} - \mu_i \boldsymbol{n}_{z, i})^T \\ (\boldsymbol{n}_{x, i} + \mu_i \boldsymbol{n}_{z, i})^T \\ (\boldsymbol{n}_{y, i} + \mu_i \boldsymbol{n}_{z, i})^T \\ \boldsymbol{n}_{z, i}^T\end{array}\right]}_{\boldsymbol{C}_i} \boldsymbol{\lambda}_i \leq \underbrace{\left[\begin{array}{c}0 \\ 0 \\ +\infty \\ +\infty \\ +\infty \end{array}\right]}_{\overline{\boldsymbol{d}}},
\end{equation}
where $\underline{\boldsymbol{d}}, \overline{\boldsymbol{d}} \in \mathbb{R}^{5}$ are the lower/upper bound vector, $\boldsymbol{C}_i \in \mathbb{R}^{5 \times3}$ is a constraint matrix at the $i$-th contact point. The first four rows of \eqref{eqn::unilateral_and_friction_cone_constraint} describe the approximated friction cone model (i.e., the friction pyramid \cite{trinkle1997dynamic}), whereas the last row encodes the unilateral constraint. 

Consider making these constraints at all contact points in a compact form, it can be expressed as:
\begin{equation}
\label{eqn::all_contact_constraints}
   \underbrace{\left[\begin{array}{c}\underline{\boldsymbol{d}} \\ \vdots \\ \underline{\boldsymbol{d}} \end{array}\right]}_{\underline{\boldsymbol{D}}} \leq \underbrace{\left[\begin{array}{ccc}\boldsymbol{C}_{1} & \cdots & \boldsymbol{0}_{1\times 3} \\ \vdots & \ddots & \vdots \\ \boldsymbol{0}_{1\times 3} & \cdots & \boldsymbol{C}_{n_c}\end{array}\right]}_{\boldsymbol{C}} \underbrace{\left[\begin{array}{c}\boldsymbol{\lambda}_1 \\ \vdots \\ \boldsymbol{\lambda}_{n_c} \end{array}\right]}_{\boldsymbol{\lambda}} \leq \underbrace{\left[\begin{array}{c}\overline{\boldsymbol{d}} \\ \vdots \\ \overline{\boldsymbol{d}} \end{array}\right]}_{\overline{\boldsymbol{D}}},
\end{equation}
with $\underline{\boldsymbol{D}}, \overline{\boldsymbol{D}} \in \mathbb{R}^{5n_c}$ are the stacked lower/upper bound vector, and $\boldsymbol{C} \in \mathbb{R}^{5n_c \times 3n_c}$ is a block diagonal matrix that includes all constraint matrices. 

\subsubsection{Unilateral and Friction Cone Constraint at Contact Point(s) for Underlying Motion}
Assume $n_m$ is the number of contact points for the motion task, and $n_f$ is the number for the force control task, such that $n_c = n_m + n_f$. As Section \ref{Sec::Motion Control for Underlying Task} mentioned, the physical constraints of the contact forces needed to implement the desired motion task for an underactuated system are considered as inequality constraints within the optimization of the constraint space error norm. According to \eqref{constraint_force}, these constraint forces $\boldsymbol{\lambda}_m \in \mathbb{R}^{3n_m}$ can be computed as:
\begin{equation}
\begin{aligned}
        \label{eqn::motion_constraint_force}
         \boldsymbol{\lambda}_m=&(\boldsymbol{J}_{c, m}^{T})^+\Big[(\boldsymbol{I}-\boldsymbol{P})[\bar{\boldsymbol{M}}(\boldsymbol{P}\boldsymbol{B}_m\boldsymbol{\tau}_{\mathrm{I}}^{*}- \boldsymbol{P h} + \dot{\boldsymbol{P}} \dot{\boldsymbol{q}}) + \boldsymbol{h}] \\
         &- (\boldsymbol{I}-\boldsymbol{P}) \boldsymbol{B}_m \boldsymbol{\tau} + (\boldsymbol{I}-\boldsymbol{P})(\bar{\boldsymbol{M}} \boldsymbol{P} - \boldsymbol{I})\boldsymbol{J}_{x}^{T} \boldsymbol{F}_x \Big],
\end{aligned}
\end{equation}
where$\boldsymbol{J}_{c, m} \in \mathbb{R}^{3n_m \times(6 + n_j)}$ represents the constraint Jacobian for the desired motion task, the motion torques $\boldsymbol{\tau}_m$ of \eqref{constraint_force} is substituted with $\boldsymbol{P}\boldsymbol{B}_m\boldsymbol{\tau}_{\mathrm{I}}^{*}$, and the constraint torques $\boldsymbol{\tau}_c$ is replaced by $(\boldsymbol{I}-\boldsymbol{P}) \boldsymbol{B}_m \boldsymbol{\tau}$ that contains the decision variable $\boldsymbol{\tau}$. Note that the $\boldsymbol{B}_m$ before $\boldsymbol{\tau}$ distributes the torque commands only to the designated joints. 

Similar to \eqref{eqn::all_contact_constraints}, the constraints of $\boldsymbol{\lambda}_m$ would be:
\begin{equation}
\label{eqn::motion_task_contact_constraints}
   \underbrace{\left[\begin{array}{c}\underline{\boldsymbol{d}} \\ \vdots \\ \underline{\boldsymbol{d}} \end{array}\right]}_{\underline{\boldsymbol{D}}_m} \leq \underbrace{\left[\begin{array}{ccc}\boldsymbol{C}_{m, 1} & \cdots & \boldsymbol{0}_{1\times 3} \\ \vdots & \ddots & \vdots \\ \boldsymbol{0}_{1\times 3} & \cdots & \boldsymbol{C}_{m, n_m}\end{array}\right]}_{\boldsymbol{C}_m} \underbrace{\left[\begin{array}{c}\boldsymbol{\lambda}_{m, 1} \\ \vdots \\ \boldsymbol{\lambda}_{m, n_m} \end{array}\right]}_{\boldsymbol{\lambda}_m} \leq \underbrace{\left[\begin{array}{c}\overline{\boldsymbol{d}} \\ \vdots \\ \overline{\boldsymbol{d}} \end{array}\right]}_{\overline{\boldsymbol{D}}_m},
\end{equation}
with $\underline{\boldsymbol{D}}_m, \overline{\boldsymbol{D}}_m \in \mathbb{R}^{5n_m}$, and $\boldsymbol{C}_m \in \mathbb{R}^{5n_m \times 3n_m}$ is a block diagonal matrix that includes the constraint matrices only for the motion task. 

\subsubsection{Unilateral and Friction Cone Constraint at Contact Point(s) for Force Control}
Akin to \eqref{eqn::motion_constraint_force}, the contact forces $\boldsymbol{\lambda}_f \in \mathbb{R}^{3n_f}$ for the force control is obtained based on $\eqref{constraint_force}$:
\begin{equation}
\begin{aligned}
        \label{eqn::force_control_constraint_force}
         \boldsymbol{\lambda}_f=&(\boldsymbol{J}_{c, f}^{T})^+\Big[(\boldsymbol{I}-\boldsymbol{P})[\bar{\boldsymbol{M}}(- \boldsymbol{P h} + \dot{\boldsymbol{P}} \dot{\boldsymbol{q}}) + \boldsymbol{h}] \\
         &- (\boldsymbol{I}-\boldsymbol{P}) \boldsymbol{B}_f \boldsymbol{\tau} + (\boldsymbol{I}-\boldsymbol{P})(\bar{\boldsymbol{M}} \boldsymbol{P} - \boldsymbol{I})\boldsymbol{J}_{x}^{T} \boldsymbol{F}_x \Big],
\end{aligned}
\end{equation}
with $\boldsymbol{J}_{c, f}\in \mathbb{R}^{3n_f \times(6 + n_j)}$ denotes the constraint Jacobian for the force control. Note that the motion torques $\boldsymbol{\tau}_m$ of \eqref{constraint_force} is set to zero vector here because the joints for the desired contact force control are not contributing to the motion control task. And constraint torques $\boldsymbol{\tau}_c$ is substituted with $(\boldsymbol{I}-\boldsymbol{P}) \boldsymbol{B}_f \boldsymbol{\tau}$ such that only the designated actuators can contribute to the force control task. 

Similarly, the physical constraints of $\boldsymbol{\lambda}_f$ would be:
\begin{equation}
\label{eqn::force_control_task_contact_constraints}
   \underbrace{\left[\begin{array}{c}\underline{\boldsymbol{d}} \\ \vdots \\ \underline{\boldsymbol{d}} \end{array}\right]}_{\underline{\boldsymbol{D}}_f} \leq \underbrace{\left[\begin{array}{ccc}\boldsymbol{C}_{f, 1} & \cdots & \boldsymbol{0}_{1\times 3} \\ \vdots & \ddots & \vdots \\ \boldsymbol{0}_{1\times 3} & \cdots & \boldsymbol{C}_{f, n_f}\end{array}\right]}_{\boldsymbol{C}_f} \underbrace{\left[\begin{array}{c}\boldsymbol{\lambda}_{f, 1} \\ \vdots \\ \boldsymbol{\lambda}_{f, n_f} \end{array}\right]}_{\boldsymbol{\lambda}_f} \leq \underbrace{\left[\begin{array}{c}\overline{\boldsymbol{d}} \\ \vdots \\ \overline{\boldsymbol{d}} \end{array}\right]}_{\overline{\boldsymbol{D}}_f},
\end{equation}
with $\underline{\boldsymbol{D}}_f, \overline{\boldsymbol{D}}_f \in \mathbb{R}^{5n_f}$, and $\boldsymbol{C}_f \in \mathbb{R}^{5n_f \times 3n_f}$ is a block diagonal matrix including the constraint matrices only for the force control task. 

\subsubsection{Torque Limits}
To make sure the sum of the optimal torques obtained from the optimization of the motion space and constraint space error norm satisfies the actuator saturation limits, the torque limit constraints can be calculated as:
\begin{equation}
    \label{eqn::torque_limit_in_constraint_space}
    \boldsymbol{\tau}_{\text{min}} - \boldsymbol{B}^T\boldsymbol{P}\boldsymbol{B}_m \boldsymbol{\tau}_{\mathrm{I}}^{*} \leq \boldsymbol{B}^T(\boldsymbol{I} - \boldsymbol{P})\boldsymbol{B}_f \boldsymbol{\tau} \leq \boldsymbol{\tau}_{\text{max}} - \boldsymbol{B}^T\boldsymbol{P}\boldsymbol{B}_m \boldsymbol{\tau}_{\mathrm{I}}^{*}.
\end{equation}

\subsubsection{Cost Function}
The optimization objective is to find the torques that can achieve the desired contact force control. Following the idea from \cite{braun2019operational}, we design the cost function as:
\begin{equation}
    \underset{\boldsymbol{\tau}}{\min }\left\|\boldsymbol{B}_f \boldsymbol{\tau}-(\boldsymbol{\tau}_{d} - \boldsymbol{P}\boldsymbol{B}_m \boldsymbol{\tau}_{\mathrm{I}}^{*})\right\|_{\boldsymbol{I}-\boldsymbol{P}}^{2}.
\end{equation}

Based on the detail of \textit{1) - 5)}, the optimization in the constraint space is summarized as:
\begin{subequations}
\begin{equation}
\label{eqn::QP_2}
    \mathrm{\textbf{QP}}_{\mathrm{\textbf{II}}}: \boldsymbol{\tau}_{\mathrm{II}}^{*}=\underset{\boldsymbol{\tau}}{\min }\left\|\boldsymbol{B}_f \boldsymbol{\tau}-(\boldsymbol{\tau}_{d} - \boldsymbol{P}\boldsymbol{B}_m \boldsymbol{\tau}_{\mathrm{I}}^{*})\right\|_{\boldsymbol{I}-\boldsymbol{P}}^{2},
\end{equation}
\begin{equation}
\label{eqn::inequality_constraints_of_QP_2}
\text{s. t. \,\,\, \eqref{eqn::motion_task_contact_constraints}, \eqref{eqn::force_control_task_contact_constraints}, \eqref{eqn::torque_limit_in_constraint_space}}. 
\end{equation}
\end{subequations}

Eventually, the ultimate controller can be defined similar to \eqref{eqn::total_torque_command_background}:
\begin{equation}
    \boldsymbol{B}\boldsymbol{\tau}^* = \boldsymbol{P}\boldsymbol{B}_m\boldsymbol{\tau}_{\mathrm{I}}^{*} + (\boldsymbol{I} - \boldsymbol{P})\boldsymbol{B}_f\boldsymbol{\tau}_{\mathrm{II}}^{*},
\end{equation}
where the first term on the right side implies the joint torques contributing to the desired motion control of the underactuated systems, while the second term only devotes to the desired contact force control. 

\section{Verification}
To verify our proposed control scheme for the contact force control, we use the quadruped robot ANYmal C in the physics engine RaiSim \cite{raisim} to conduct a task, that is, using its front left leg to exerting desired contact force to the ground while its other three legs support the free-floating torso to track a constant pose (i.e., $x = y = 0 \,m$ and $z = 0.57 \,m$ for position, and $roll = pitch = yaw = 0 \,rad$ for orientation). This task aims to imitate an underactuated quadrupedal robot pressing an object by applying a desired force at the contact point while keeping standing. Specifically, we design two reference force profiles: sinewave and step force, for performance verification. For the sinewave force, we set $F_x = 30\cdot \text{sin}(0.2t) \, N$, $F_y = 20\cdot \text{sin}(t) \, N$ and $F_z = 140 - 50\cdot \text{sin}(2t) \, N$ with different frequency and magnitude. For the step force, we set $F_x$ and $F_y$ always to $0 \, N$, and $F_z$ starts from $100 \, N$, jumps to $130 \, N$, then to $160 \, N$ and back to $130 \, N$, and finalizes at $100 \, N$. These force reference profiles are assumed to be feasible and are also used to imitate the commands from a user.

We compare the force control performance of three controllers labeled as:
\begin{itemize}
    \item \textit{proposed:} Our proposed control scheme \eqref{eqn::QP_1} - \eqref{eqn::inequality_constraints_of_QP_2}.
    \item \textit{HOWSM:} Hierarchical optimization structure \eqref{eqn::QP_1} and \eqref{eqn::QP_2}, but without using the selection matrices design to decouple the force and motion task (i.e., replacing $\boldsymbol{B}_m$ in \eqref{eqn::QP_1} and $\boldsymbol{B}_f$ in \eqref{eqn::QP_2} by $\boldsymbol{B}$).
    \item \textit{PIDCWCU:} Projected inverse dynamics contact wrench control \eqref{eqn::Contact-consistent motion generation and contact wrench control for the underactuation} for underactuated robots \cite{dehio2018modeling}.
\end{itemize}

\begin{figure*}[t!]
            \centering
            \includegraphics[width=0.8975\textwidth]{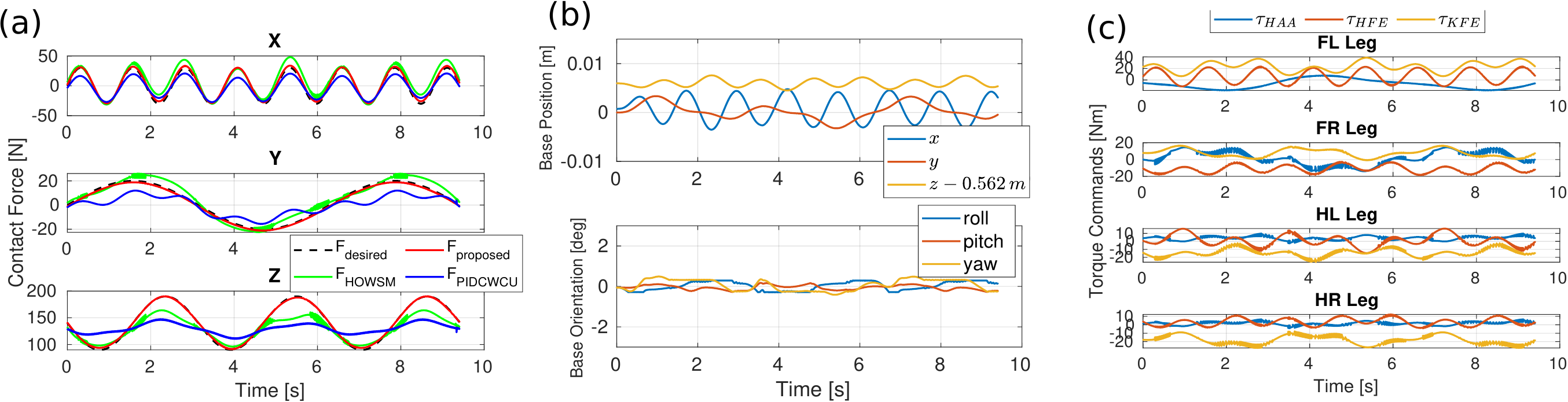}
            \caption{Simulation results of sinewave force tracking. (a) Front left foot's contact force tracking the performance of three controllers; (b) and (c): Base motion and optimized torque inputs when using our proposed controller.}
            \label{fig:simulation_sinwave_tracking}
\end{figure*}
\begin{figure*}[t!]
            \centering
            \includegraphics[width=0.8975\textwidth]{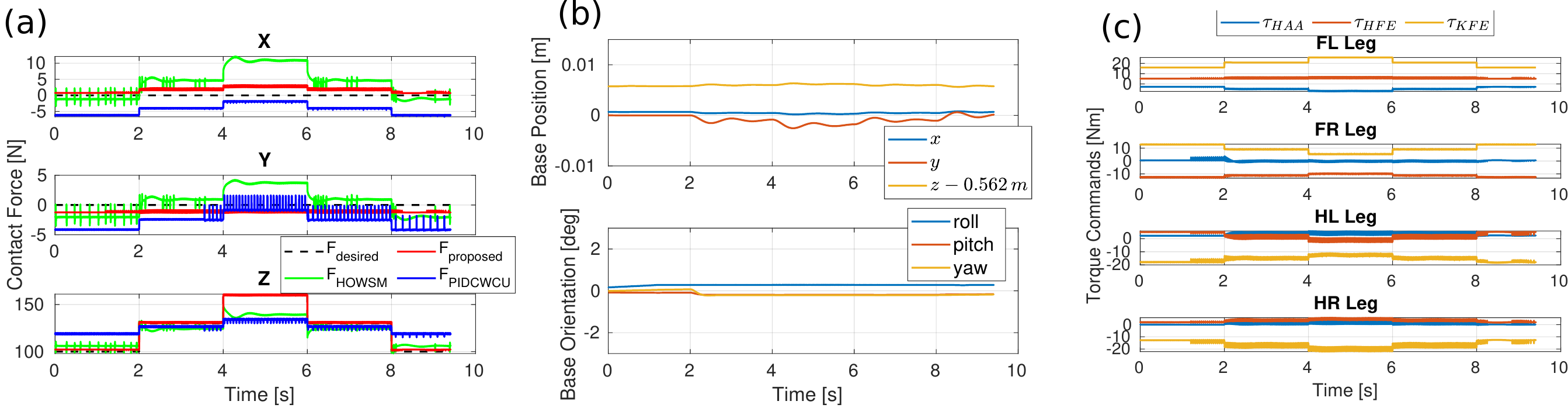}
            \caption{Simulation results of step force tracking. (a) Front left foot's contact force tracking the performance of three controllers; (b) and (c): Base motion and optimized torque inputs when using our proposed controller.}
            \label{fig:simulation_step_force_tracking}
\end{figure*}

To be fair, the same control gains $\boldsymbol{K}_{p} = 2000 \cdot \boldsymbol{I}_6$, $\boldsymbol{K}_{d} = 100 \cdot \boldsymbol{I}_6$ are used for all cases. 
Figs. \ref{fig:simulation_sinwave_tracking} and \ref{fig:simulation_step_force_tracking} illustrate the simulation results of sinewave and step force tracking, respectively. We observe that our proposed controller demonstrated more accurate contact force control performance in any direction, although slight tracking errors exist. It reveals that our controller is a reactive controller that has a fast response to user-defined contact force without using any force planning. The force error in our method is mainly induced by two reasons: 1) we formulate the constraint force tracking as a cost function in the constraint space, and therefore the constraint force control could be traded-off when the inequality constraints must be enforced; 2) We do not use the Jacobians at the exact contact points to construct the constraint Jacobian $\boldsymbol{J}_c$, as it is unfeasible in hardware experiment. To simulate as much as possible the situation in hardware, we instead use the Jacobians in the centers of the contact feet for $\boldsymbol{J}_c$, and thus it causes the force error. Moreover, our controller induces minimal base motion, as shown in (b) of Fig. \ref{fig:simulation_sinwave_tracking} and \ref{fig:simulation_step_force_tracking}.

The purpose of comparing the force control performance with \textit{HOWSM} is to emphasize that using two selection matrices in our controller is the key to decouple the force and motion task towards the exact contact force control. Without using them, the motion and force control task are still highly coupled because the contact foot used for the force control will also contribute to the motion task via the original selection matrix $\boldsymbol{B}$. For \textit{PIDCWCU}, the comparison results show that this controller cannot approach the exact contact force control, as the additional term $(\boldsymbol{I - P})\boldsymbol{\tau}_u$ that is designed for resolving underactuation would induce a disturbance to the contact force control and thus fail the force tracking task.


\section{Conclusion and Future Work}
This paper presents a novel control scheme that aims to approach the exact interaction force control for underactuated quadrupedal systems. Based on PIDC, the constraint forces can be formulated in the function of actuation torques, allowing us to design the desired contact forces. To resolve underactuation, we follow the idea of constrained hierarchical optimization that solves two QP optimization problems in the motion and constraint space sequentially and satisfies the physical constraints at the same time. To decouple the motion and force control task and achieve the exact contact force control to the greatest extent, we propose a strategy of applying two selection matrices to the cost function of the QP design. Combining the hierarchical structure and the strategy, a novel control scheme is proposed. Compared to the constrained hierarchical optimization method without using two selection matrices and the PIDC method for contact force control, our controller performed more precise contact force control since the coupling effect between motion and force control was greatly 
mitigated, facilitating the exact interaction force control for underactuated quadrupedal systems. In the future, we will implement our algorithm on the hardware, and implement it to other underactuated quadruped systems like quadruped mobile manipulators. Moreover, we will further decouple the motion and contact force control task in an analytical way.

\bibliographystyle{./IEEEtran} 
\bibliography{./IEEEabrv,./mybib}

\begin{thebibliography}{10}
\providecommand{\url}[1]{#1}
\csname url@rmstyle\endcsname
\providecommand{\newblock}{\relax}
\providecommand{\bibinfo}[2]{#2}
\providecommand\BIBentrySTDinterwordspacing{\spaceskip=0pt\relax}
\providecommand\BIBentryALTinterwordstretchfactor{4}
\providecommand\BIBentryALTinterwordspacing{\spaceskip=\fontdimen2\font plus
\BIBentryALTinterwordstretchfactor\fontdimen3\font minus
  \fontdimen4\font\relax}
\providecommand\BIBforeignlanguage[2]{{%
\expandafter\ifx\csname l@#1\endcsname\relax
\typeout{** WARNING: IEEEtran.bst: No hyphenation pattern has been}%
\typeout{** loaded for the language `#1'. Using the pattern for}%
\typeout{** the default language instead.}%
\else
\language=\csname l@#1\endcsname
\fi
#2}}

\bibitem{XinOneLeg2020}
G.~Xin, J.~Smith, D.~Rytz, W.~Wolfslag, H.-C. Lin, and M.~Mistry, ``Bounded
  haptic teleoperation of a quadruped robot’s foot posture for sensing and
  manipulation,'' in \emph{2020 IEEE International Conference on Robotics and
  Automation (ICRA)}, 2020, pp. 1431--1437.

\bibitem{Aghili2005}
F.~Aghili, ``A unified approach for inverse and direct dynamics of constrained
  multibody systems based on linear projection operator: applications to
  control and simulation,'' \emph{IEEE Transactions on Robotics}, vol.~21,
  no.~5, pp. 834--849, 2005.

\bibitem{Mistry2010Floating}
M.~Mistry, J.~Buchli, and S.~Schaal, ``Inverse dynamics control of floating
  base systems using orthogonal decomposition,'' in \emph{2010 IEEE
  International Conference on Robotics and Automation}, 2010, pp. 3406--3412.

\bibitem{mistry2012operational}
M.~Mistry and L.~Righetti, ``Operational space control of constrained and
  underactuated systems,'' in \emph{Robotics: Science and systems}, vol.~7,
  2012, pp. 225--232.

\bibitem{chu2021operational}
X.~Chu, Y.~Tang, A.~M. Giordano, T.~Chen, and K.~S. Au, ``Operational space
  control for planar pa n--1 underactuated manipulators using orthogonal
  projection and quadratic programming,'' in \emph{2021 IEEE International
  Conference on Robotics and Automation (ICRA)}.\hskip 1em plus 0.5em minus
  0.4em\relax IEEE, 2021, pp. 12\,853--12\,859.

\bibitem{xin2020optimization}
G.~Xin, W.~Wolfslag, H.-C. Lin, C.~Tiseo, and M.~Mistry, ``An
  optimization-based locomotion controller for quadruped robots leveraging
  cartesian impedance control,'' \emph{Frontiers in Robotics and AI}, vol.~7,
  p.~48, 2020.

\bibitem{dehio2018modeling}
N.~Dehio, J.~Smith, D.~L. Wigand, G.~Xin, H.-C. Lin, J.~J. Steil, and
  M.~Mistry, ``Modeling and control of multi-arm and multi-leg robots:
  Compensating for object dynamics during grasping,'' in \emph{2018 IEEE
  International Conference on Robotics and Automation (ICRA)}.\hskip 1em plus
  0.5em minus 0.4em\relax IEEE, 2018, pp. 294--301.

\bibitem{dehio2022enabling}
N.~Dehio, J.~Smith, D.~L. Wigand, P.~Mohammadi, M.~Mistry, and J.~J. Steil,
  ``Enabling impedance-based physical human--multi--robot collaboration:
  Experiments with four torque-controlled manipulators,'' \emph{The
  International Journal of Robotics Research}, vol.~41, no.~1, pp. 68--84,
  2022.

\bibitem{RighettiDistribution2013}
L.~Righetti, J.~Buchli, M.~Mistry, M.~Kalakrishnan, and S.~Schaal, ``Optimal
  distribution of contact forces with inverse-dynamics control,'' \emph{The
  International Journal of Robotics Research}, vol.~32, no.~3, pp. 280--298,
  2013.

\bibitem{aghili2005unified}
F.~Aghili, ``A unified approach for inverse and direct dynamics of constrained
  multibody systems based on linear projection operator: applications to
  control and simulation,'' \emph{IEEE Transactions on Robotics}, vol.~21,
  no.~5, pp. 834--849, 2005.

\bibitem{ott2008cartesian}
C.~Ott, \emph{Cartesian impedance control of redundant and flexible-joint
  robots}.\hskip 1em plus 0.5em minus 0.4em\relax Springer, 2008.

\bibitem{ortenzi2014experimental}
V.~Ortenzi, M.~Adjigble, J.~A. Kuo, R.~Stolkin, and M.~Mistry, ``An
  experimental study of robot control during environmental contacts based on
  projected operational space dynamics,'' in \emph{2014 IEEE-RAS International
  Conference on Humanoid Robots}.\hskip 1em plus 0.5em minus 0.4em\relax IEEE,
  2014, pp. 407--412.

\bibitem{ortenzi2015projected}
V.~Ortenzi, R.~Stolkin, J.~A. Kuo, and M.~Mistry, ``Projected inverse dynamics
  control and optimal control for robots in contact with the environment: A
  comparison,'' in \emph{2015 IEEE/RSJ International Conference on Intelligent
  Robots and Systems (IROS)}.\hskip 1em plus 0.5em minus 0.4em\relax IEEE,
  2015, pp. 4009--4015.

\bibitem{braun2019operational}
D.~J. Braun, Y.~Chen, and L.~Li, ``Operational space control under actuation
  constraints using strictly convex optimization,'' \emph{IEEE Transactions on
  Robotics}, vol.~36, no.~1, pp. 302--309, 2019.

\bibitem{lin2018projected}
H.-C. Lin, J.~Smith, K.~K. Babarahmati, N.~Dehio, and M.~Mistry, ``A projected
  inverse dynamics approach for multi-arm cartesian impedance control,'' in
  \emph{2018 IEEE International Conference on Robotics and Automation
  (ICRA)}.\hskip 1em plus 0.5em minus 0.4em\relax IEEE, 2018, pp. 5421--5428.

\bibitem{trinkle1997dynamic}
J.~C. Trinkle, J.-S. Pang, S.~Sudarsky, and G.~Lo, ``On dynamic
  multi-rigid-body contact problems with coulomb friction,'' \emph{ZAMM-Journal
  of Applied Mathematics and Mechanics/Zeitschrift f{\"u}r Angewandte
  Mathematik und Mechanik}, vol.~77, no.~4, pp. 267--279, 1997.

\bibitem{raisim}
\BIBentryALTinterwordspacing
J.~Hwangbo, J.~Lee, and M.~Hutter, ``Per-contact iteration method for solving
  contact dynamics,'' \emph{IEEE Robotics and Automation Letters}, vol.~3,
  no.~2, pp. 895--902, 2018. [Online]. Available: \url{www.raisim.com}
\BIBentrySTDinterwordspacing

\end{thebibliography}

\end{document}